\documentclass[final]{cvpr}

\usepackage{times}
\usepackage{epsfig}
\usepackage{graphicx}
\usepackage{amsmath}
\usepackage{amssymb}
\usepackage{acro}
\usepackage{xspace}
\usepackage{multirow}
\usepackage{booktabs}
\usepackage{makecell}

\usepackage[pagebackref=true,breaklinks=true,colorlinks,bookmarks=false]{hyperref}

\DeclareAcronym{FPN}{
short=FPN,
long=feature pyramid network
}
\DeclareAcronym{PAN}{
short=PAN,
long=path aggregation network
}
\DeclareAcronym{NMS}{
short=NMS,
long=non-maximum suppression
}
\DeclareAcronym{GSI}{
short=GSI,
long=Gaussian-smoothed interpolation
}
\DeclareAcronym{FPs}{
short=FPs,
long=false positives
}
\DeclareAcronym{FNs}{
short=FNs,
long=false negatives
}
\DeclareAcronym{IDs}{
short=IDs,
long=ID switches
}

\def\cvprPaperID{311} 
\def\confYear{CVPR 2021}

\begin{document}

\title{PieTrack: An MOT solution based on synthetic data training and self-supervised domain adaptation}

\author{Yirui Wang \and Shenghua He \and Youbao Tang \and Jingyu Chen \and Honghao Zhou \and Sanliang Hong \and Junjie Liang \and Yanxin Huang \and Ning Zhang \and Ruei-Sung Lin  \and Mei Han \\
PAII Inc. \& Ping An Bank Co., Ltd.\\
\and
}

\maketitle

\begin{abstract}
In order to cope with the increasing demand for labeling data and privacy issues with human detection, synthetic data has been used as a substitute and showing promising results in human detection and tracking tasks. We participate in the 7th Workshop on Benchmarking Multi-Target Tracking (BMTT), themed on "How Far Can Synthetic Data Take us"? Our solution, \textbf{PieTrack}, is developed based on synthetic data without using any pre-trained weights. We propose a self-supervised domain adaptation method that enables mitigating the domain shift issue between the synthetic (\eg, \emph{MOTSynth}) and real data (\eg, \emph{MOT17}) without involving extra human labels. By leveraging the proposed multi-scale ensemble inference, we achieved a final HOTA score of \textbf{58.7} on the MOT17 testing set, ranked third place in the challenge.

\end{abstract}

\section{Introduction}
Multiple object tracking (MOT) is an active research area with profound applications. With more advanced and bigger deep learning-based models, the need of labeled real data has also been increased greatly. This is especially the case in human-centered applications such as detection, tracking, action recognition and spatial-temporal localization, etc. Although labeling people in clear background with large-scale bounding-box is not difficult, precising labelling of people in small-scale with occlusion is not easy and could be error-prone. Meanwhile, there has been an increasing awareness of the privacy protection and avoid exposing real human faces for deep learning applications. This is such an important and ethical precaution but difficult in practice due to the face obfuscation (blurring, mosaicing, etc.) still far from practical in most human detection-related applications \cite{Kaiyu}.

On the other hand, the synthetic data has served as a good alternative in certain applications such as autonomous-driven vehicles \cite{Hu3DT19}, with virtual driving datasets such as virtual KITTI \cite{gaidon2016virtual}, GTA5\cite{richter2016playing}, VIPER \cite{richter2017playing}. Part of the success with the virtual driving datasets is due to the rigid body of cars which does not have flexible limbs like human being.

Last year (2021), thanks to the advanced game engine with physics simulation as well as human body simulation, MOTSynth, a large-scale synthetic data has been released for the multiple object tracking \cite{fabbri2021motsynth}. This dataset benefits the community in multiple areas of research and applications including pedestrian detection, tracking, re-identification, and instance segmentation.
Such a synthetic data drew our attention because of its advantages in both labeling effort and privacy preservation. In particular, we are delighted to be part of the 7th BMTT Challenges ``MOTSynth-MOT-CVPR22'', and our participating approach has the following highlights:
\begin{itemize}
\item[--] A two-stage detection and association strategy is adopted, where our detector is trained using synthetic data from scratch without pre-trained weights.
\item[--] Self-supervised domain adaptation approach is used to fill the gap between the synthetic trained detector and the real MOT data, by using \textbf{only} unlabeled MOT17 training split.
\item[--] Multi-scale ensemble inference by taking different resolutions as input is used. This strategy helps to further boost the performance.
\end{itemize}

\section{Related Work}
\textbf{Tracking-by-Detection.} Powerful detection capability thanks to the deep-learning models, enables better tracking performance in the MOT domain. One-stage approaches \cite{tracktor_2019_ICCV, zhou2020tracking,peng2020ctracker}, tackle the detection and the tracking tasks together in an end-to-end fashion and achieves high-speed process. However, this group of approaches suffers a lower performance than the two-stage because of its compromise between detection and tracking branches of the deep learning model.  Recent developments of YOLO-based approaches \cite{redmon2018yolov3, bochkovskiy2020yolov4, ge2021ota} provides high-speed, high-accuracy detectors.  These advances facilitate the two-stage approaches by striking a balance between the detection speed and the data association \cite{zhang2021bytetrack}.

\textbf{Synthetic Training Data.} Autonomous driving applications have adopted synthetic data in its model training since early time. Virtual KITTI \cite{gaidon2016virtual} cloned the real data to virtual proxy for detection validation. Omni-MOT \cite{sun2020simultaneous} provided a dataset, focuses on car-based detection and tracking.  Parallel Domain is a commercial content provider for MOT of cars \cite{tokmakov2021learning}. In the people-centered MOT domain, JTA (Joint Track Auto) \cite{fabbri2018learning} proposed a dataset on CG human poses for estimation and tracking. MOTSynth \cite{fabbri2021motsynth} is the most comprehensive dataset for pedestrian-based MOT and have drawn a lot attention since its release.
We are also drawn to this dataset and participated the Challenge using the MOTSynth. Next section presents our methodology for the challenge  with detailed explanation and illustration.

\section{Method}
\begin{figure}[t]
    \centering
    \noindent\includegraphics[width=\linewidth]{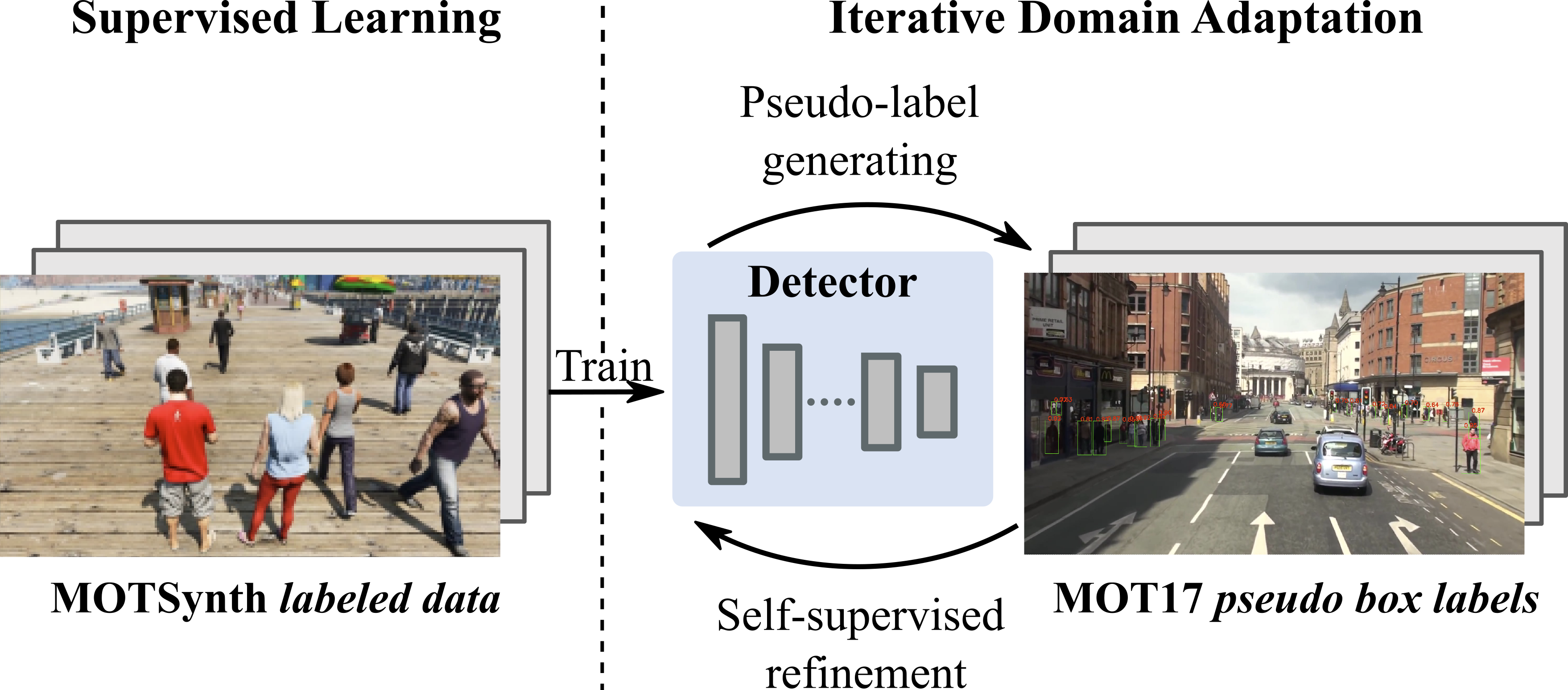}
    \caption{\textbf{Overview of the iterative domain adaptation pipeline}. Firstly, the detector is trained on pure MOTSynth dataset. Then we perform the iterative domain adaptation by repeatedly generating and training with the pseudo bounding-box labels of the MOT17 training set.}
    \label{fig:DA}
\end{figure}

In this work, we adopt the tracking by detection paradigm and propose an iterative approach to alleviate the domain gap between the synthetic and the real data. To follow the competition rule that \textbf{no human labels for the MOT17 dataset can be used for fine-tuning}, we propose a self-supervised pseudo-label mining and refining procedure on the MOT17 training set to shrink the domain gap. And we also adopt an multi-scale ensemble strategy to improve the robustness during the inference. The overview of the training framework is shown in Fig. \ref{fig:DA}.
\subsection{Multi-Object Detection and Tracking}
\textbf{Object Detection Network.} The proposed PieTrack employs YOLOX \cite{ge2021yolox} detection network to detect object of interests (\eg, pedestrian) in each frame. The network is built upon YOLOv3 \cite{redmon2018yolov3} architecture with modified CSPNet \cite{wang2020cspnet} backbone and PAN \cite{liu2018path} head to enhance gradient propagation and feature hierarchy. YOLOX decouples the classification and regression tasks into separate branches as they are prone to conflict with each other \cite{song2020revisiting,wu2020rethinking}. The detector adopts an anchor-free design to avoid domain-specific anchor configuration and reduce computational cost during inference \cite{ge2021yolox}. To alleviate the imbalance between positive and negative samples, YOLOX consider each $3 \times 3$ area around the true object centers as positive samples. Moreover, a simplified OTA \cite{ge2021ota} strategy, named SimOTA, is proposed to improve the training efficiency without sacrificing the performance. The network can be customized to different sizes and trained in an end-to-end manner. In our work, we adopt YOLOX-X as our detection network and we follow \cite{ge2021yolox} to train the detector from scratch \textbf{without using any pretrained weights}.

\textbf{Data Association and Tracking.} We adopt a two-stage data association method proposed in \cite{zhang2021bytetrack} with modifications to partially mitigate the domain shift issue. In the first stage, the association is performed between the high confidence detection boxes and all tracklets. In the second stage, the association is further performed between the unmatches less confident detection boxes and the remain tracklets. Different from the original strategy in \cite{zhang2021bytetrack} that excludes lost tracklets from the second association, we propose to consider both tracked and lost tracklets for sensitivity consideration. Because of domain shift issue, we observed significantly increased false negatives. By considering the lost tracklets in both stage, we empirically find it effectively increases the tracking sensitivity by at least $1.1\%$. In addition, we employ NSA Kalman algorithm \cite{du2021giaotracker} as the dynamic motion model and use \ac{GSI} to remediate missing tracklets \cite{du2022strongsort}. 

\subsection{Self-supervised Domain Adaptation}\label{sec.ssl}
\textbf{Motivations.} Domain shift issue is the most challenging part in this task. It is noticeable that patterns, such as pedestrian's appearance, lighting conditions, distractors, \etc{}, are significantly different between the source domain (\ie{} MOTSynth) and the target domain (\ie{} MOT17), casting a shadow on achieving high detection/tracking performance on the testing scenario. Therefore, in our work, we focus on leveraging self-supervised domain adaptation method to boost the detection performance by using \textbf{only} unlabeled MOT17 training split.

\textbf{Domain Adaptation via Iterative Refinement}
Domain adaptation is crucial for acquiring good generalization ability on unseen data, especially when there is a domain gap between the training and testing scenario. Recently, training from synthetic data draws many attentions, as it is an efficient way to collect large-scale dataset. In our work, we propose to leverage a simple yet effective self-supervised iterative refinement approach to address the domain shift issue. We firstly train the detector for $80$ epochs on the whole MOTSynth dataset (with a subsampling ratio of 10), and we validate the performance on the MOT17 validation half to determine the best model. Then, we start the iterative domain adaptation by firstly using a pre-defined detection confidence threshold ($T=0.5$ for the initial iteration, and $T=0.1$ for the rest) to generate pseudo bounding-box labels for the MOT17 training set, and we fine-tune the detector with the pseudo-labels for another $N$ ($N=40$ for the initial iteration, and $N=20$ for the rest) epochs. We repeat this process until there is no improvement when validating the model with MOT17 training set true labels. We find this process significantly reduce false negatives and noticeably improve the overall tracking performance, as demonstrated in the Sec. \ref{sec.exp}.

\subsection{Multi-scale Ensemble Inference}
To improve the robustness of detection results during inference, similar to \cite{casado2020ensemble,cheng2021scalable,hsieh2021automated}, we employ an ensemble strategy to further boost the model performance. We take three different resolutions as inputs (\eg, $640 \times 1152$, $800 \times 1440$, and $1280 \times 2304$), and augment each input by horizontal flipping. This creates three pairs of inputs for the detection network. We design a strategy to make the input resolution compatible with the predicted box size. To be specific, we only allow the lowest resolution input to predict boxes that larger than $32 \times 32$, and restrict the largest input to produce boxes that smaller than $96 \times 96$. All box predictions from the medium resolution inputs are used.  And we collect all the legal predictions together and perform \ac{NMS} at once. The effectiveness of this multi-scale ensemble strategy is demonstrated in Sec. \ref{sec.exp}.

\section{Experiments}\label{sec.exp}
\subsection{Dataset and Evaluation Metrics}
MOTSynth \cite{fabbri2021motsynth} is a large-scale synthetic dataset for object detection and tracking. It consists of $768$ sequences and results in $1,382$k frames. We employ MOTSynth as the only source of labeled data for the development. For parameter tuning and methodology validation, we use MOT17 \cite{milan2016mot16} training split as our validation set. We primarily use HOTA \cite{luiten2021hota} and MOTA \cite{mota} to assess the overall tracking performance, and also take \ac{FPs}, \ac{FNs}, and \ac{IDs} as references.  

\subsection{Ablation Study}
We conduct ablation study to reveal the effectiveness of the proposed iterative domain adaptation and multi-scale ensemble inference. As demonstrated in Table~\ref{tab:ablation}, the baseline method that trained on the MOTSynth labeled dataset produces large amount of false negatives when testing on the MOT17 training set. We hypotheses that this poor performance stems from the domain shift issue as described in Sec.~\ref{sec.ssl}. Therefore, we further conduct iterative domain adaptation to improve the sensitivity of the model. As can be seen, we repeat this process twice and obtain consistent and significant improvements over the baseline method. By introducing the multi-scale ensemble inference, we further boost the HOTA score to $57.68$.

\begin{table}[tb]
  \centering
  \caption{\textbf{Ablation study on the iterative domain adaptation and the multi-scale ensemble learning on MOT17 training set}.} 
  \begin{tabular}{lp{1.cm}<{\centering}p{1.cm}<{\centering}p{1.cm}<{\centering}p{1.cm}<{\centering}p{1.cm}<{\centering}}
  \toprule
  Method & HOTA & MOTA & FP & FN & IDs \\ \midrule
  Baseline     & 52.87  & 60.50  & \textbf{10553}  & 33258 & 514     \\ \midrule
  Iter.1     & 55.66  & 63.9  & 11426  & 28643 & 517     \\
  Iter.2     & 56.82  & 63.8  & 11574  & 28569 & \textbf{456}     \\ \midrule
  Ensemble     & \textbf{57.68}  & \textbf{64.30}  & 15528  & \textbf{23985} & 534     \\
  \bottomrule
  \end{tabular}
  \label{tab:ablation} \vspace{-2mm}
\end{table}

\subsection{MOTSynth Challenge Performance}
We use the best model obtained by the iterative domain adaptation to perform inference the MOT17 testing split. We submit the results to the MOTChallenge platform, which yields a HOTA score of \textbf{58.7}. 

\section{Conclusion}
We present our approach for 7th BMTT's Challenge on MOTSynth-MOT-CVPR22. We adopt tracking-by-detection paradigm with two-stage approach, where detector is trained from scratch using MOTSynth data. Domain adaptation uses self-supervised fashion where only the unlabeled MOT17 training data is used. We also further boosts our performance with multiple-scale ensemble inference. We achieved a HOTA score of \textbf{58.7} as the final ranking result showing on the challenge website.

{\small
\bibliographystyle{ieee_fullname}
\bibliography{cvpr}

\begin{thebibliography}{10}\itemsep=-1pt

\bibitem{tracktor_2019_ICCV}
Philipp Bergmann, Tim Meinhardt, and Laura Leal{-}Taix{\'{e}}.
\newblock Tracking without bells and whistles.
\newblock In {\em The IEEE International Conference on Computer Vision (ICCV)},
  October 2019.

\bibitem{mota}
Keni Bernardin and Rainer Stiefelhagen.
\newblock Evaluating multiple object tracking performance: the clear mot
  metrics.
\newblock {\em EURASIP Journal on Image and Video Processing}, 2008:1--10,
  2008.

\bibitem{bochkovskiy2020yolov4}
Alexey Bochkovskiy, Chien-Yao Wang, and Hong-Yuan~Mark Liao.
\newblock Yolov4: Optimal speed and accuracy of object detection.
\newblock {\em arXiv preprint arXiv:2004.10934}, 2020.

\bibitem{casado2020ensemble}
{\'A}ngela Casado-Garc{\'\i}a and J{\'o}nathan Heras.
\newblock Ensemble methods for object detection.
\newblock In {\em ECAI 2020}, pages 2688--2695. IOS Press, 2020.

\bibitem{cheng2021scalable}
Chi-Tung Cheng, Yirui Wang, Huan-Wu Chen, Po-Meng Hsiao, Chun-Nan Yeh, Chi-Hsun
  Hsieh, Shun Miao, Jing Xiao, Chien-Hung Liao, and Le Lu.
\newblock A scalable physician-level deep learning algorithm detects universal
  trauma on pelvic radiographs.
\newblock {\em Nature communications}, 12(1):1--10, 2021.

\bibitem{du2022strongsort}
Yunhao Du, Yang Song, Bo Yang, and Yanyun Zhao.
\newblock Strongsort: Make deepsort great again.
\newblock {\em arXiv preprint arXiv:2202.13514}, 2022.

\bibitem{du2021giaotracker}
Yunhao Du, Junfeng Wan, Yanyun Zhao, Binyu Zhang, Zhihang Tong, and Junhao
  Dong.
\newblock Giaotracker: A comprehensive framework for mcmot with global
  information and optimizing strategies in visdrone 2021.
\newblock In {\em Proceedings of the IEEE/CVF International Conference on
  Computer Vision}, pages 2809--2819, 2021.

\bibitem{fabbri2021motsynth}
Matteo Fabbri, Guillem Bras{\'o}, Gianluca Maugeri, Orcun Cetintas, Riccardo
  Gasparini, Aljo{\v{s}}a O{\v{s}}ep, Simone Calderara, Laura Leal-Taix{\'e},
  and Rita Cucchiara.
\newblock Motsynth: How can synthetic data help pedestrian detection and
  tracking?
\newblock In {\em Proceedings of the IEEE/CVF International Conference on
  Computer Vision}, pages 10849--10859, 2021.

\bibitem{fabbri2018learning}
Matteo Fabbri, Fabio Lanzi, Simone Calderara, Andrea Palazzi, Roberto Vezzani,
  and Rita Cucchiara.
\newblock Learning to detect and track visible and occluded body joints in a
  virtual world.
\newblock In {\em Proceedings of the European conference on computer vision
  (ECCV)}, pages 430--446, 2018.

\bibitem{gaidon2016virtual}
Adrien Gaidon, Qiao Wang, Yohann Cabon, and Eleonora Vig.
\newblock Virtual worlds as proxy for multi-object tracking analysis.
\newblock In {\em Proceedings of the IEEE conference on computer vision and
  pattern recognition}, pages 4340--4349, 2016.

\bibitem{ge2021ota}
Zheng Ge, Songtao Liu, Zeming Li, Osamu Yoshie, and Jian Sun.
\newblock Ota: Optimal transport assignment for object detection.
\newblock In {\em Proceedings of the IEEE/CVF Conference on Computer Vision and
  Pattern Recognition}, pages 303--312, 2021.

\bibitem{ge2021yolox}
Zheng Ge, Songtao Liu, Feng Wang, Zeming Li, and Jian Sun.
\newblock Yolox: Exceeding yolo series in 2021.
\newblock {\em arXiv preprint arXiv:2107.08430}, 2021.

\bibitem{hsieh2021automated}
Chen-I Hsieh, Kang Zheng, Chihung Lin, Ling Mei, Le Lu, Weijian Li, Fang-Ping
  Chen, Yirui Wang, Xiaoyun Zhou, Fakai Wang, et~al.
\newblock Automated bone mineral density prediction and fracture risk
  assessment using plain radiographs via deep learning.
\newblock {\em Nature communications}, 12(1):1--9, 2021.

\bibitem{Hu3DT19}
Hou-Ning Hu, Qi-Zhi Cai, Dequan Wang, Ji Lin, Min Sun, Philipp Krähenbühl,
  Trevor Darrell, and Fisher Yu.
\newblock Joint monocular 3d vehicle detection and tracking.
\newblock 2019.

\bibitem{liu2018path}
Shu Liu, Lu Qi, Haifang Qin, Jianping Shi, and Jiaya Jia.
\newblock Path aggregation network for instance segmentation.
\newblock In {\em Proceedings of the IEEE conference on computer vision and
  pattern recognition}, pages 8759--8768, 2018.

\bibitem{luiten2021hota}
Jonathon Luiten, Aljosa Osep, Patrick Dendorfer, Philip Torr, Andreas Geiger,
  Laura Leal-Taix{\'e}, and Bastian Leibe.
\newblock Hota: A higher order metric for evaluating multi-object tracking.
\newblock {\em International journal of computer vision}, 129(2):548--578,
  2021.

\bibitem{milan2016mot16}
Anton Milan, Laura Leal-Taix{\'e}, Ian Reid, Stefan Roth, and Konrad Schindler.
\newblock Mot16: A benchmark for multi-object tracking.
\newblock {\em arXiv preprint arXiv:1603.00831}, 2016.

\bibitem{peng2020ctracker}
Jinlong Peng, Changan Wang, Fangbin Wan, Yang Wu, Yabiao Wang, Ying Tai,
  Chengjie Wang, Jilin Li, Feiyue Huang, and Yanwei Fu.
\newblock Chained-tracker: Chaining paired attentive regression results for
  end-to-end joint multiple-object detection and tracking.
\newblock In {\em Proceedings of the European Conference on Computer Vision},
  2020.

\bibitem{redmon2018yolov3}
Joseph Redmon and Ali Farhadi.
\newblock Yolov3: An incremental improvement.
\newblock {\em arXiv preprint arXiv:1804.02767}, 2018.

\bibitem{richter2017playing}
Stephan~R Richter, Zeeshan Hayder, and Vladlen Koltun.
\newblock Playing for benchmarks.
\newblock In {\em Proceedings of the IEEE International Conference on Computer
  Vision}, pages 2213--2222, 2017.

\bibitem{richter2016playing}
Stephan~R Richter, Vibhav Vineet, Stefan Roth, and Vladlen Koltun.
\newblock Playing for data: Ground truth from computer games.
\newblock In {\em European conference on computer vision}, pages 102--118.
  Springer, 2016.

\bibitem{song2020revisiting}
Guanglu Song, Yu Liu, and Xiaogang Wang.
\newblock Revisiting the sibling head in object detector.
\newblock In {\em Proceedings of the IEEE/CVF Conference on Computer Vision and
  Pattern Recognition}, pages 11563--11572, 2020.

\bibitem{sun2020simultaneous}
ShiJie Sun, Naveed Akhtar, XiangYu Song, HuanSheng Song, Ajmal Mian, and
  Mubarak Shah.
\newblock Simultaneous detection and tracking with motion modelling for
  multiple object tracking.
\newblock In {\em European Conference on Computer Vision}, pages 626--643.
  Springer, 2020.

\bibitem{tokmakov2021learning}
Pavel Tokmakov, Jie Li, Wolfram Burgard, and Adrien Gaidon.
\newblock Learning to track with object permanence.
\newblock In {\em Proceedings of the IEEE/CVF International Conference on
  Computer Vision}, pages 10860--10869, 2021.

\bibitem{wang2020cspnet}
Chien-Yao Wang, Hong-Yuan~Mark Liao, Yueh-Hua Wu, Ping-Yang Chen, Jun-Wei
  Hsieh, and I-Hau Yeh.
\newblock Cspnet: A new backbone that can enhance learning capability of cnn.
\newblock In {\em Proceedings of the IEEE/CVF conference on computer vision and
  pattern recognition workshops}, pages 390--391, 2020.

\bibitem{wu2020rethinking}
Yue Wu, Yinpeng Chen, Lu Yuan, Zicheng Liu, Lijuan Wang, Hongzhi Li, and Yun
  Fu.
\newblock Rethinking classification and localization for object detection.
\newblock In {\em Proceedings of the IEEE/CVF conference on computer vision and
  pattern recognition}, pages 10186--10195, 2020.

\bibitem{Kaiyu}
Kaiyu Yang, Jacqueline Yau, Li Fei{-}Fei, Jia Deng, and Olga Russakovsky.
\newblock A study of face obfuscation in imagenet.
\newblock {\em arXiv preprint arXiv:2103.06191}, 2021.

\bibitem{zhang2021bytetrack}
Yifu Zhang, Peize Sun, Yi Jiang, Dongdong Yu, Zehuan Yuan, Ping Luo, Wenyu Liu,
  and Xinggang Wang.
\newblock Bytetrack: Multi-object tracking by associating every detection box.
\newblock {\em arXiv preprint arXiv:2110.06864}, 2021.

\bibitem{zhou2020tracking}
Xingyi Zhou, Vladlen Koltun, and Philipp Kr{\"a}henb{\"u}hl.
\newblock Tracking objects as points.
\newblock {\em ECCV}, 2020.

\end{thebibliography}
}

\end{document}